\documentclass[runningheads]{llncs}

\usepackage{eccv}

\usepackage{eccvabbrv}

\usepackage{graphicx}
\usepackage{booktabs}
\usepackage{multirow}
\usepackage{wrapfig}
\usepackage{marvosym}
\usepackage[accsupp]{axessibility}  %

\usepackage{multibib}

\newcites{supp}{Supplementary References}

\usepackage{hyperref}

\usepackage{orcidlink}

\begin{document}

\title{Task-Agnostic Incremental Vision-Language Object Detection via Prompt Augmentation and Distribution-Aware Fusion}

\titlerunning{TA-IVLOD via Prompt Augmentation and Distribution-Aware Fusion}

\author{Yonghan Jiang\inst{1}\orcidlink{0009-0002-8196-9771} \and
Zhengyuan Xie\inst{1}\orcidlink{0000-0003-2522-285X} \and
Wenchu Liu\inst{1}\orcidlink{0009-0002-2607-1221} \and
Linlan Huang\inst{1}\orcidlink{0009-0007-0515-1202} \and
Fei Yang\inst{1,3}\orcidlink{0000-0003-4099-6511} \and
Xialei Liu\inst{1,2,3}\orcidlink{0000-0001-8534-3026}\thanks{Corresponding author.}%
}

\authorrunning{Y.~Jiang et al.}

\institute{VCIP, CS, Nankai University, Tianjin, China\and
AAIS, Nankai University, Tianjin, China \and
NKIARI, Shenzhen Futian, Shenzhen, China
\\
\email{\{yonghanjiang, xiezhengyuan, wenchuliu, huanglinlan\}@mail.nankai.edu.cn}\\
\email{\{feiyang, xialei\}@nankai.edu.cn}
}

\maketitle

\begin{abstract}
Incremental Vision-Language Object Detection (IVLOD) empowers pre-trained open-vocabulary detectors to continuously learn new visual concepts from sequential tasks without forgetting foundational knowledge. However, existing methods predominantly rely on oracle task identities during inference, severely limiting their open-world practicality. In this paper, we formalize Task-Agnostic IVLOD (TA-IVLOD), a more realistic setting where such priors are unavailable, requiring the model to simultaneously recognize classes from all learned tasks. This unconstrained setting exposes severe performance degradation due to cross-task semantic interference and parameter conflicts. To tackle these challenges, we propose TADA, a novel modular framework. Specifically, TADA introduces Stochastic Prompt Augmentation to mitigate semantic interference via training-time noise injection, and Test-Time Distribution-Aware Fusion to dynamically weight class-specific LoRA experts, effectively resolving parameter conflicts. Extensive evaluations demonstrate that our method significantly outperforms baselines in standard IVLOD and yields substantial improvements in the rigorous TA-IVLOD setting on the ODinW-13 benchmark, while effectively preserving zero-shot generalization on the MS COCO dataset. The code is available at \url{https://github.com/yonghanjiang/TADA}.
\end{abstract}

\section{Introduction}
\label{sec:intro}

Object detection~\cite{carion2020end,zhang2022dino} has long been a fundamental task in computer vision. Recently, Vision-Language Object Detection (VLOD), powered by large-scale
vision-language pre-training~\cite{liu2024grounding}, has enabled Open-Vocabulary Object Detection
(OVOD)~\cite{gu2021open,minderer2023scaling,maaz2022class,li2022grounded,zhang2022glipv2}, allowing detectors to localize objects beyond
predefined categories through textual queries. Despite their strong zero-shot
generalization, these models often suffer from substantial performance degradation
when deployed in specialized downstream domains whose visual distributions differ
from web-scale pre-training data.

To bridge this gap, Incremental Vision-Language Object Detection (IVLOD) has been introduced to continuously adapt OVOD models to novel tasks while mitigating catastrophic forgetting~\cite{deng2024zero}. To achieve efficient continual learning, recent vision-language frameworks have explored various mechanisms to preserve prior knowledge~\cite{zhang2023vqacl,chen2023continual,yu2024select,liu2025c,huang2025mind,huang2024class}. A common paradigm involves decoupling the acquisition of new concepts from the preservation of old capabilities, often by isolating task-specific parameters within the network. Such structural modifications allow for domain-specific updates without altering the core knowledge stored in the backbone, ensuring that the model remains robust on previously learned tasks while accommodating new information.

While current modular adaptation strategies (e.g., ZiRa~\cite{deng2024zero} and DitHub~\cite{cappellino2025dithub}) successfully balance plasticity and stability, they fundamentally rely on a highly restrictive assumption: the Task-Incremental setting~\cite{oren2021defense,liu2020multi}. By depending on oracle task information during inference, these methods assume the task identity of each test sample is provided a priori. Consequently, they evaluate the model using a constrained prompt comprising only the classes from that specific task, simplifying the problem into processing isolated, single-task prompts. This reliance severely limits their applicability in open-world scenarios where task boundaries are unknown, forcing the model to confront an overwhelming semantic space of all encountered classes simultaneously.

\begin{wrapfigure}[19]{r}{0.433\textwidth}
  \centering
    \includegraphics[width=0.433\textwidth]{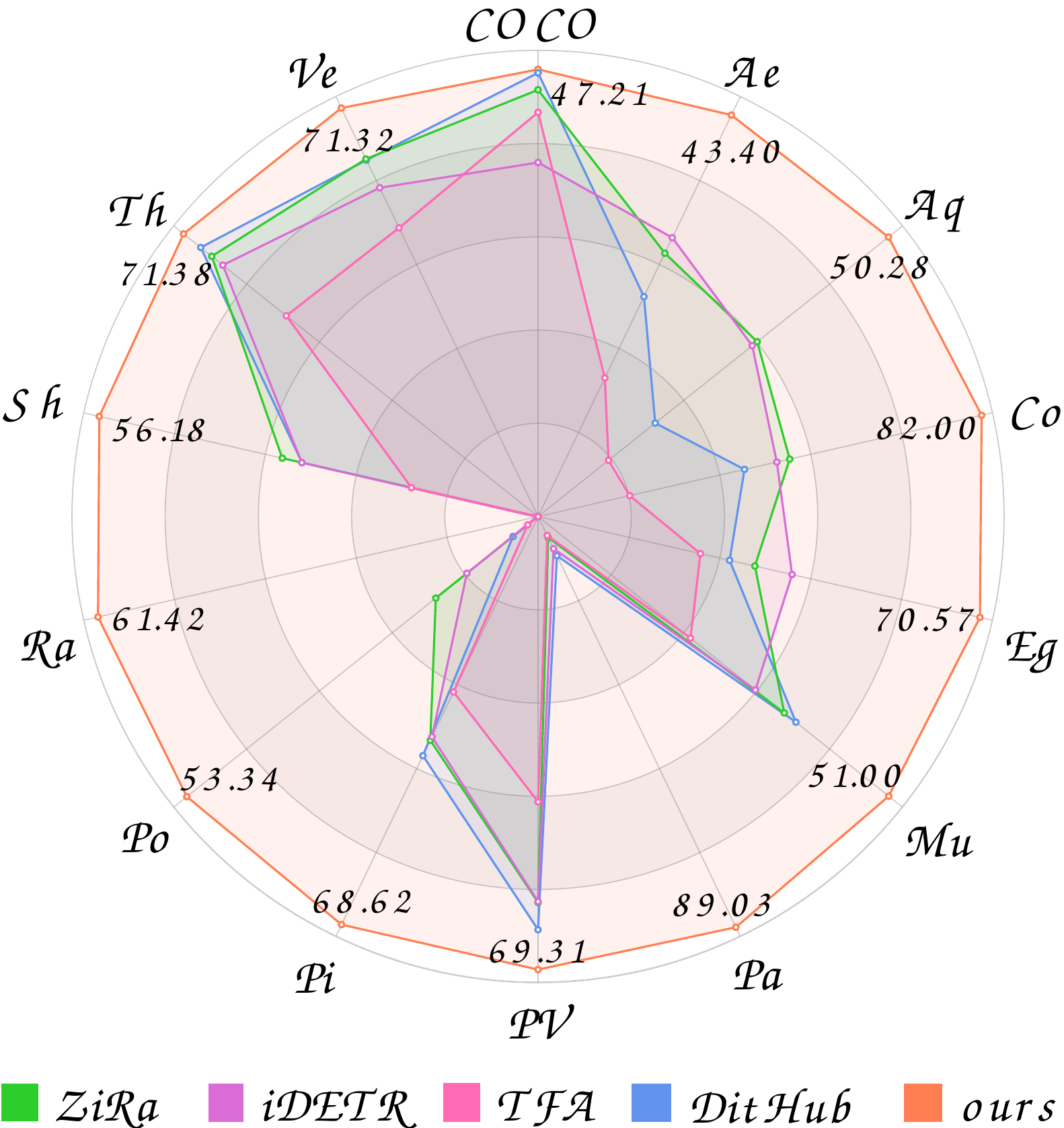}
  \caption{Our model achieves superior performance on 14 datasets under 
the TA-IVLOD setting.}
\label{fig:radar}
\end{wrapfigure}

\begin{figure}[t]
  \centering
    \begin{subfigure}{\linewidth}
    \centering
    \includegraphics[width=0.9\linewidth]{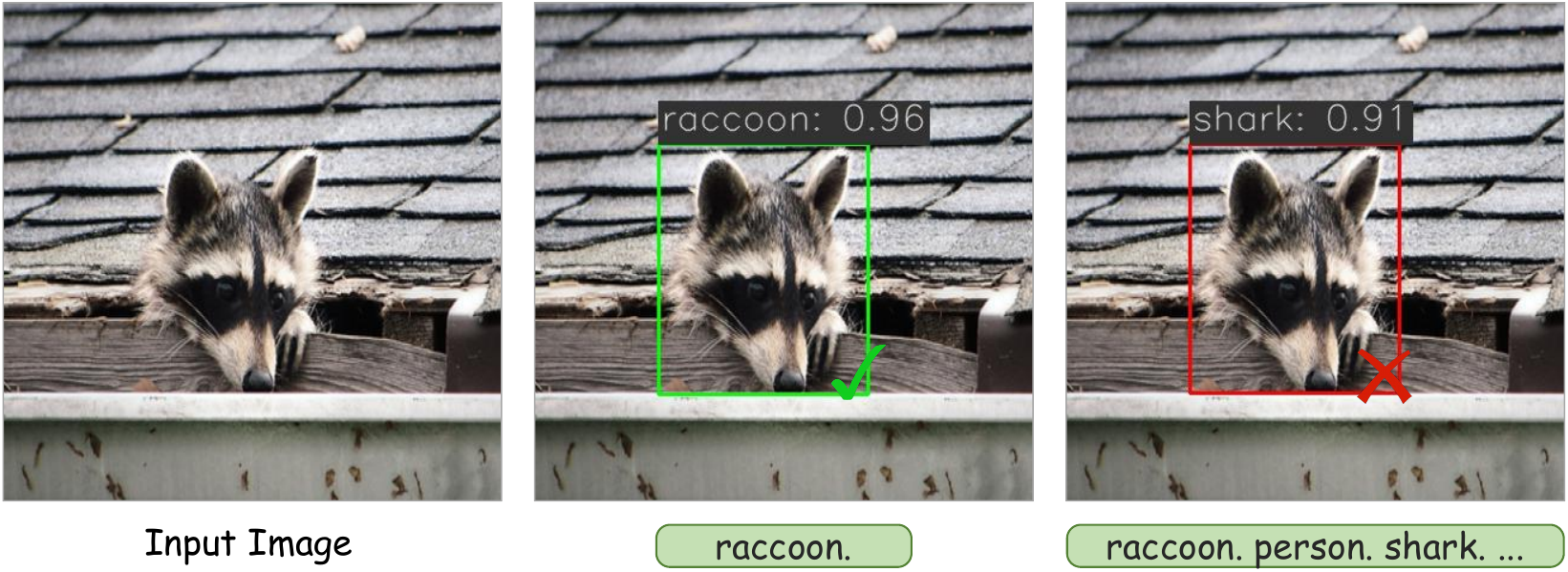}
  \end{subfigure}\hfill
\caption{Comparison of detection results. (Middle) Single-task prompt in the standard IVLOD setting. (Right) Cross-task prompt in the TA-IVLOD setting.}
  \label{fig:iv_ta_comp}
\end{figure}

To align with the demands of real-world applications, we formally introduce a more realistic and challenging setting: Task-Agnostic Incremental Vision-Language Object Detection (TA-IVLOD). Under this setting, the model operates entirely without task identity priors during inference. However, this paradigm shift exposes a fundamental bottleneck we term Prompt Interference. Since mainstream OVOD architectures rely on deep cross-modal fusion, transitioning from single-task prompts to cross-task prompts naturally causes severe attention dilution. Crucially, the sequential nature of continual learning severely exacerbates this vulnerability. During training, the model optimizes cross-modal alignments for disjoint subsets of classes in complete isolation. Consequently, when these independently learned concepts are concatenated into a cross-task prompt during task-agnostic inference, the visual attention allocated to different semantic classes inevitably collides. Lacking the ability to coordinate attention across concurrently presented, heterogeneous classes, this semantic conflict directly translates into parameter conflict and catastrophic performance degradation, as illustrated in \cref{fig:iv_ta_comp}.

To address these challenges, we present TADA (Task-Agnostic Distribution-Aware Adaptation), a modular framework for task-agnostic inference. Rather than relying on naive parameter merging or constrained training setups, our approach addresses two key challenges: prompt interference and parameter conflict. Specifically, we introduce Stochastic Prompt Augmentation (SPA) to proactively simulate semantic noise during optimization, and propose Test-Time Distribution-Aware Fusion (TTDF) to dynamically align visual feature distributions at inference. By synergizing these two modules, our framework enables the detector to maintain robust cross-modal alignments. As visually summarized in \cref{fig:radar}, our method delivers superior detection performance in the TA-IVLOD setting.

The main contributions of this paper are summarized as follows:
\begin{itemize}
    \item We introduce TA-IVLOD, a realistic and challenging setting devoid of task oracle priors.
    \item We design Stochastic Prompt Augmentation to mitigate cross-task semantic interference via training-time noise injection.
    \item We propose Test-Time Distribution-Aware Fusion to dynamically weight class-specific experts, effectively resolving parameter conflicts.
    \item Extensive experiments on the ODinW-13 benchmark demonstrate substantial performance gains in TA-IVLOD and consistent improvements in standard IVLOD.
\end{itemize}

\section{Related Work}

\noindent\textbf{Incremental Vision-Language Object Detection.} 
Vision-Language Object Detection (VLOD)~\cite{zhu2024survey} aims to generalize detection to unseen classes via cross-modal alignment, effectively overcoming the bottleneck of fixed category sets. In this domain, pioneering works such as ViLD~\cite{gu2021open} distill knowledge from CLIP~\cite{radford2021learning} to align region proposals with text embeddings. Subsequently, GLIP~\cite{li2022grounded} reformulates detection as phrase grounding for deeper fusion, and Grounding DINO~\cite{liu2024grounding} leverages Transformer architectures to significantly boost open-set accuracy. However, these models primarily assume a static vocabulary, struggling to adapt to dynamic data streams. 

To address this, Incremental Vision-Language Object Detection (IVLOD)~\cite{deng2024zero} requires models to sequentially learn new classes while mitigating catastrophic forgetting of base classes and preserving zero-shot generalization. This rigorous tri-objective optimization distinguishes IVLOD from conventional incremental learning, which typically neglects open-world capabilities. Recent methods tackle this via modular adaptation. ZiRa~\cite{deng2024zero} incorporates zero-interference loss and re-parameterization, while DitHub~\cite{cappellino2025dithub} proposes a scalable LoRA-based framework for flexible class extension. However, their reliance on task-specific oracle prompts during inference severely restricts their applicability in unconstrained environments. To overcome this limitation, we formalize TA-IVLOD, a challenging new paradigm designed for true open-world scenarios where task boundaries are entirely unknown.

\noindent\textbf{Parameter Fusion and LoRA Integration.}
Recent works~\cite{liu2024dora,zhou2022conditional,li2024clip,jia2022visual,peng2025parameter} have explored parameter-efficient fine-tuning (PEFT) methods, such as Adapter Tuning~\cite{houlsby2019parameter} and Prompt Learning~\cite{zhou2022learning,zhou2022conditional}, to adapt pre-trained models without updating all parameters. Among these, Low-Rank Adaptation (LoRA)~\cite{hu2022lora} has gained widespread adoption due to its competitive efficiency. However, maintaining separate LoRA modules in incremental learning leads to linear growth in inference latency and memory consumption. To mitigate this, LoRA fusion techniques~\cite{zhou2022conditional,zheng2025decouple,huang2023lorahub,wu2024mixture,yadav2023ties} merge multiple low-rank matrices into a single adapter via weight averaging~\cite{asadi2025does} or orthogonal projection~\cite{zheng2025decouple}. 

In the context of IVLOD, DitHub~\cite{cappellino2025dithub} adopts this paradigm by statically averaging the $\mathbf{A}$ matrices of LoRA adapters across observed classes while sharing a common $\mathbf{B}$ matrix. While such static averaging preserves general knowledge within isolated, single-task environments, it inevitably triggers severe \textbf{parameter conflicts} when processing the unified, cross-task prompts required for task-agnostic inference. In contrast, our approach dynamically weights class-specific parameters by aligning them with the input visual distribution at test time. This distribution-aware strategy effectively resolves parameter conflicts and ensures robust cross-modal alignment without requiring oracle task identities.
\section{Task-Agnostic IVLOD}

\subsection{Problem Formulation}
Following the standard incremental learning paradigm, we formulate the training process as a sequence of $N$ tasks, denoted as $\mathcal{T} = \{T_1, T_2, \dots, T_N\}$. Each task $T_k$ is associated with a specific label set $\mathcal{C}_k$. The goal is to sequentially learn these tasks while mitigating catastrophic forgetting.

Existing IVLOD approaches predominantly adopt an evaluation protocol that implicitly aligns with the Task-Incremental setting. Under this scenario, task information is implicitly exploited during prompt construction. In particular, for a test image belonging to task $T_k$, the query prompt $\mathcal{P}_{I}$ is restricted strictly to the classes within that specific task: $\mathcal{P}_{I}^{(k)} = \{c \mid c \in \mathcal{C}_k\}$. This implicit task prior simplifies the problem by partitioning the global label space into disjoint subspaces, thereby artificially circumventing the interference from classes across different tasks.

To align with realistic open-world scenarios, we introduce the TA-IVLOD setting. In this challenging setup, no task priors are provided during inference. Consequently, instead of utilizing constrained, task-specific queries, the model is required to use a unified query prompt encompassing the union of all learned classes: $\mathcal{P}_{TA} = \bigcup_{i=1}^N \mathcal{C}_i$. Furthermore, unlike the standard IVLOD protocol that evaluates each task $i$ independently on its corresponding isolated test set $\mathcal{D}_{test}^{(i)}$, TA-IVLOD mandates a joint evaluation over a comprehensive global test set formed by uniting all encountered task domains: $\mathcal{D}_{test}^{TA} = \bigcup_{i=1}^N \mathcal{D}_{test}^{(i)}$. Consequently, the model must localize and classify objects within a significantly expanded semantic candidate space in a single inference pass. This evaluation protocol introduces a high volume of negative samples for each task—specifically images devoid of its target classes—thereby increasing the risk of cross-task false positives and providing a more rigorous assessment of open-world robustness.

\begin{figure}[!t]
  \centering
  \begin{subfigure}{0.24\linewidth}
    \includegraphics[width=\linewidth]{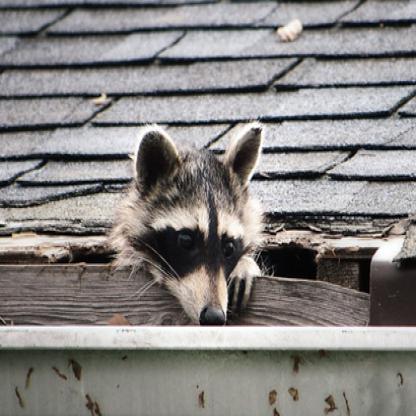}
    \caption{Image}
    \label{subfig:image}
  \end{subfigure}\hfill
  \begin{subfigure}{0.24\linewidth}
    \includegraphics[width=\linewidth]{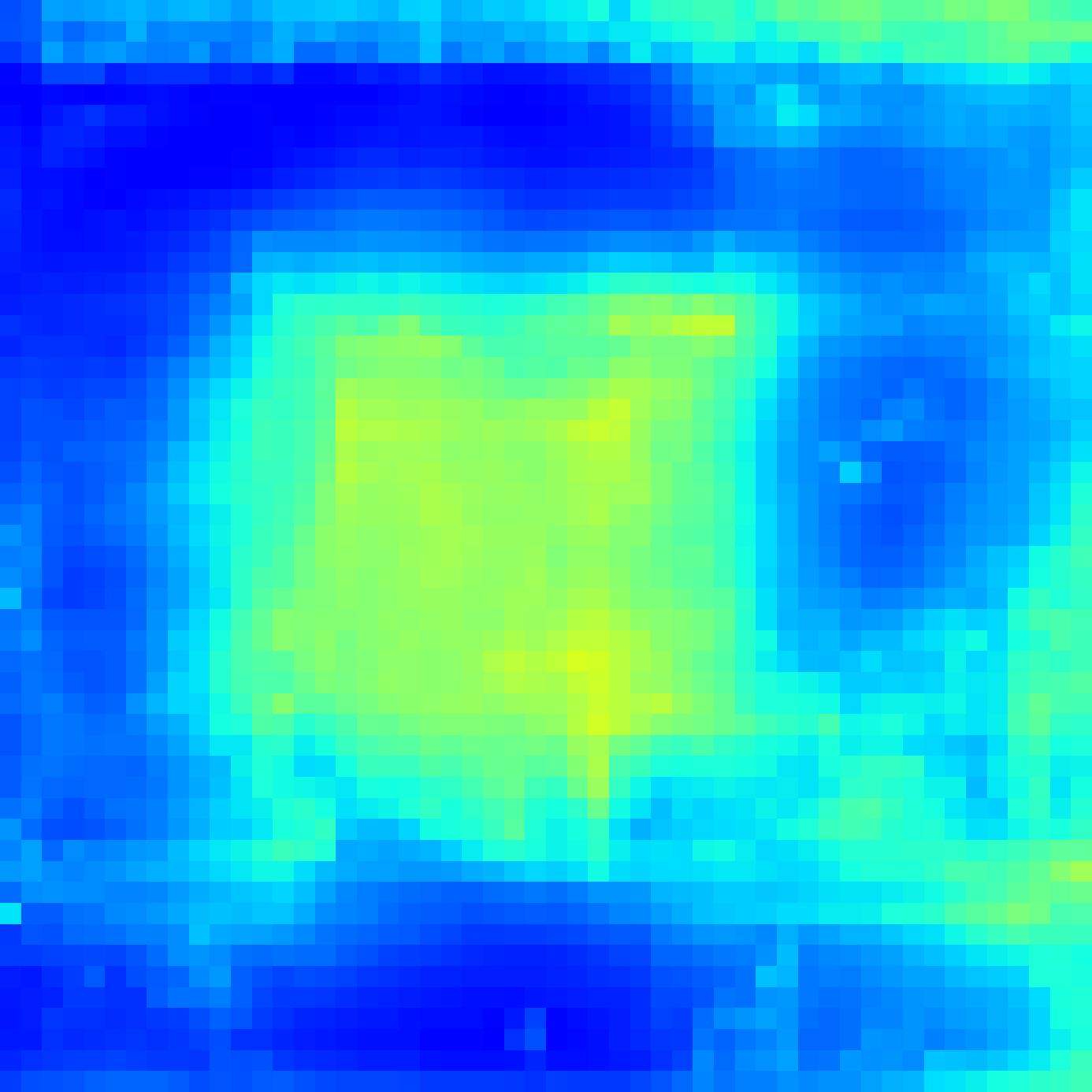}
    \caption{``raccoon.''}
    \label{subfig:simple_attn}
  \end{subfigure}\hfill
  \begin{subfigure}{0.5\linewidth}
    \includegraphics[width=0.48\linewidth]{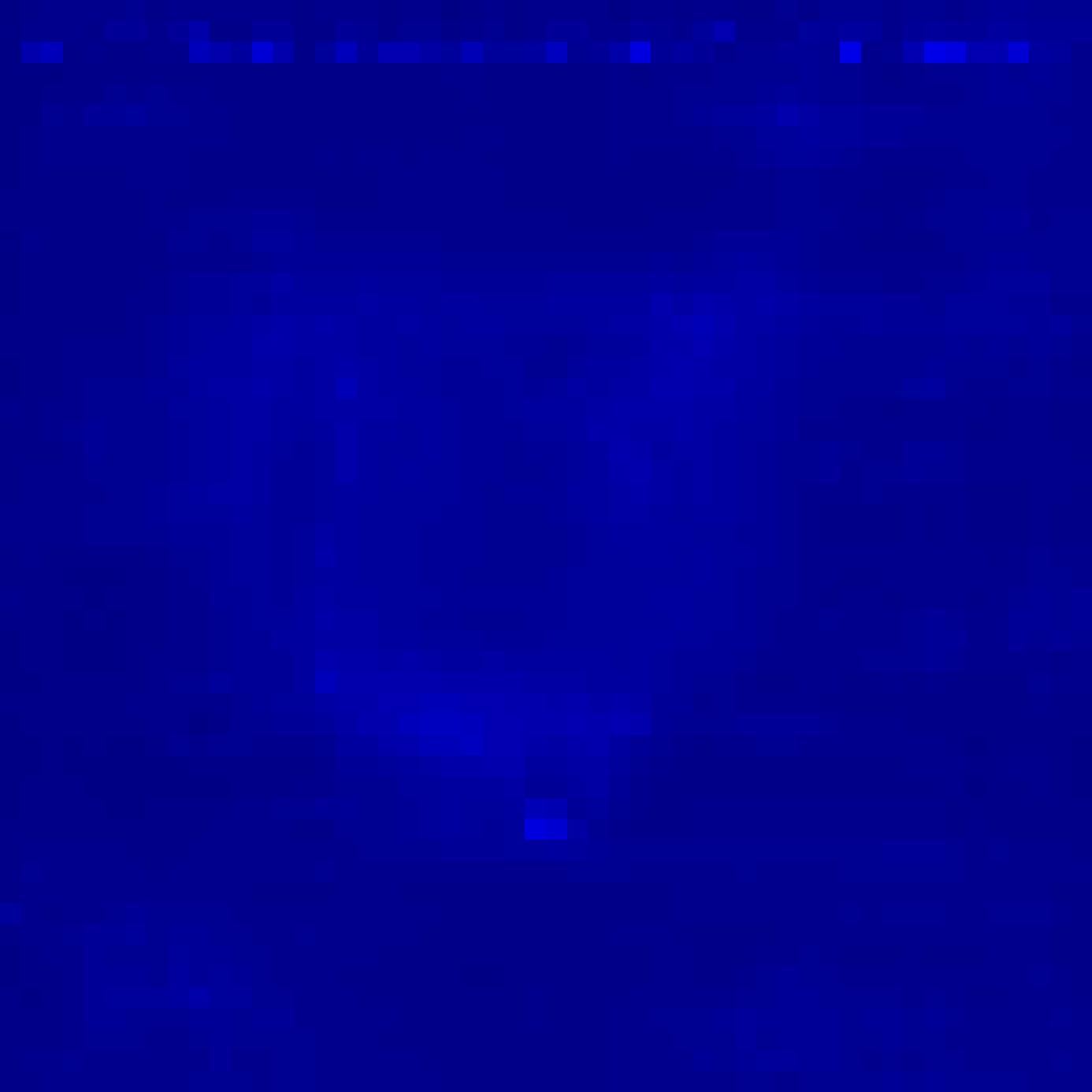}
    \includegraphics[width=0.48\linewidth]{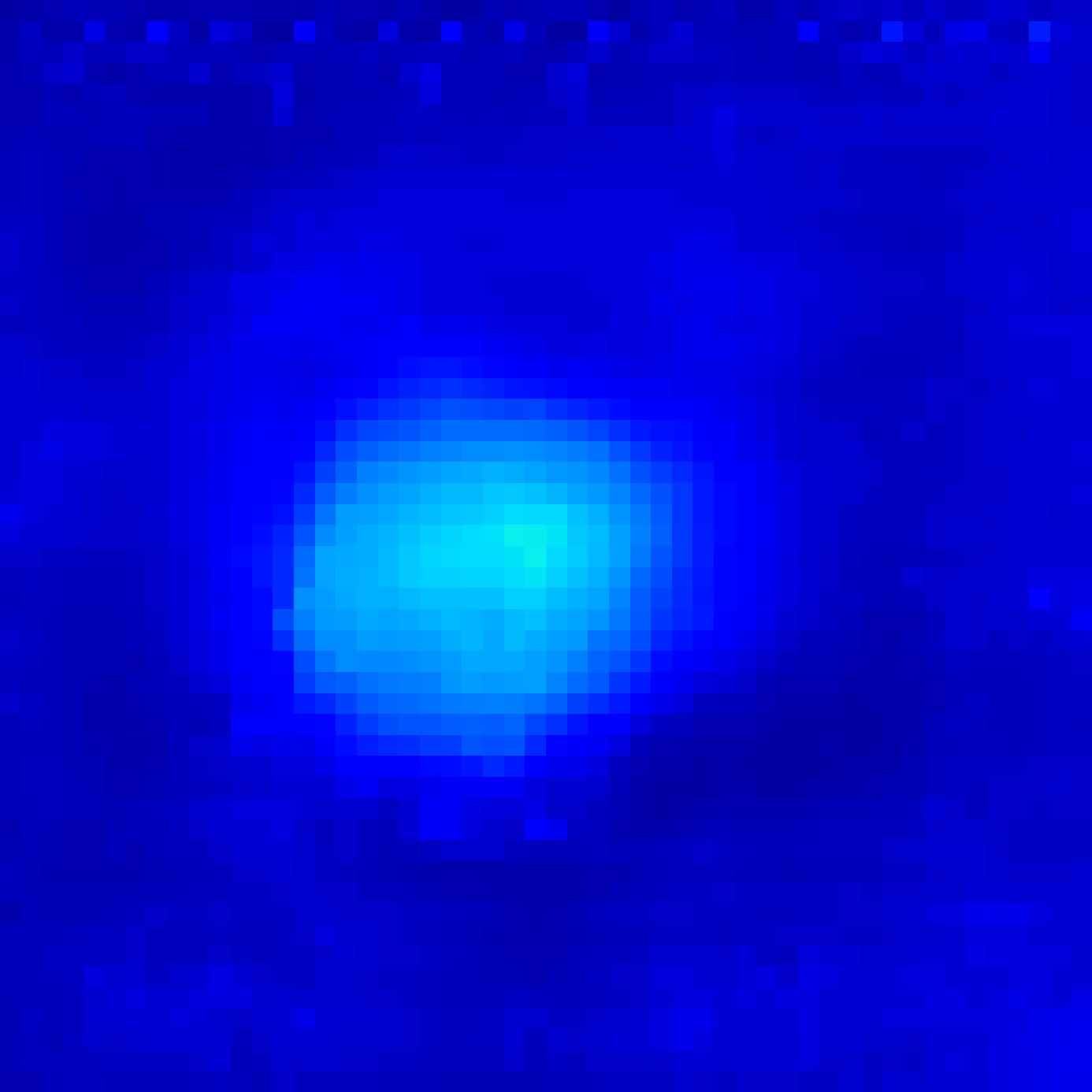}
    \caption{``raccoon. person. shark. ...''}
    \label{subfig:cross_attn}
  \end{subfigure}
\caption{Visualization of image feature attention towards the correct target class (``raccoon''). (a) Original image. (b) Attention under a single-task prompt. (c) Attention under a cross-task prompt. While the baseline (left) fails to focus on the correct class due to semantic interference, our proposed method (right) accurately attends to the target object.}
  \label{fig:attention}
\end{figure}

\begin{figure}[!t]
    \centering
    \includegraphics[width=\textwidth]{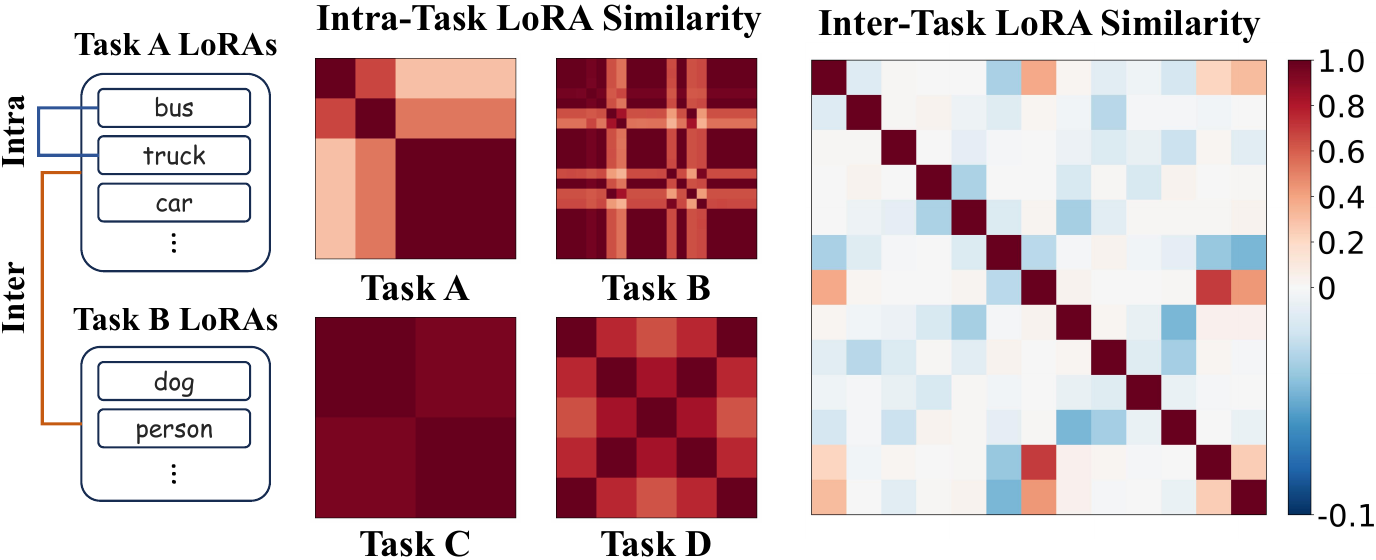}
    \caption{Cosine similarity matrices of class-specific LoRA experts. The left panels show the intra-task similarity among LoRA modules learned for different classes within the same task. The right panel shows the inter-task similarity matrix across all 13 tasks, which is computed by averaging the class-specific LoRA modules within each task beforehand.}
    \label{fig:lora_sim}
\end{figure}

\subsection{Challenges}
\label{sec:ta_challenge}
\noindent \textbf{Prompt Interference.}
As incremental tasks accumulate, the target class set continuously expands. To ensure real-world practicality, the model must perform unified inference across all learned classes simultaneously, without relying on task oracle priors. This fundamentally differs from the sparse, task-specific prompts used during training, creating a severe training-inference prompt discrepancy. In OVOD architectures, the text prompt serves as an implicit conditional adapter that dictates the alignment of visual features. Subjecting the model to an extensive, consolidated inference prompt triggers Prompt Interference, where the cluttered textual context tends to dilute the attention weights assigned to key semantic tokens. This attention dilution and contextual misalignment introduce substantial semantic noise, preventing the visual features from accurately attending to their corresponding target tokens. To visualize this, we extract the image-to-text attention weights corresponding to the ground-truth class (\eg, ``raccoon'') from the final cross-attention layer of the Grounding DINO feature enhancer. As shown in \cref{fig:attention}, single-task prompts yield focused and well-aligned attention (\cref{subfig:simple_attn}), whereas cross-task prompts cause severe attention diffusion and misalignment (\cref{subfig:cross_attn}, left). This demonstrates that prompt expansion acts as a strong distractor, overwhelming the target semantics and degrading visual-textual alignment.

\noindent \textbf{Parameter Conflict.}
Crucially, the sequential nature of continual learning severely amplifies the aforementioned prompt-induced interference. Because the model is strictly optimized on disjoint tasks with task-local objectives and without access to data from other tasks, the learning process inherently lacks cross-task coordination. This vulnerability is particularly evident in representative modular frameworks, such as DitHub, which allocate a distinct LoRA expert for each individual class. Driven by the isolated training paradigm, these independently learned experts inevitably follow divergent optimization trajectories. When these uncoordinated representations are forced into a unified semantic space to process cross-task prompts during task-agnostic inference, they suffer from intrinsic structural conflicts and pronounced representational misalignment. To empirically validate this, we visualize the cosine similarity among these class-specific LoRA experts. Prior studies~\cite{wu2025sd} establish that merging LoRA modules with low directional similarity significantly exacerbates performance degradation. As illustrated in \cref{fig:lora_sim}, experts optimized within the same task maintain high correlation (left panels), ensuring that intra-task fusion introduces minimal parameter conflict. Conversely, the inter-task similarity across different tasks is significantly lower (right panel). Consequently, fusing these highly divergent modules under the unconstrained TA-IVLOD setting leads to severe parameter conflicts, exposing a fundamental weakness of existing incremental adaptation schemes.

\section{Method}

\begin{figure}[t]
    \centering
    \includegraphics[width=\textwidth]{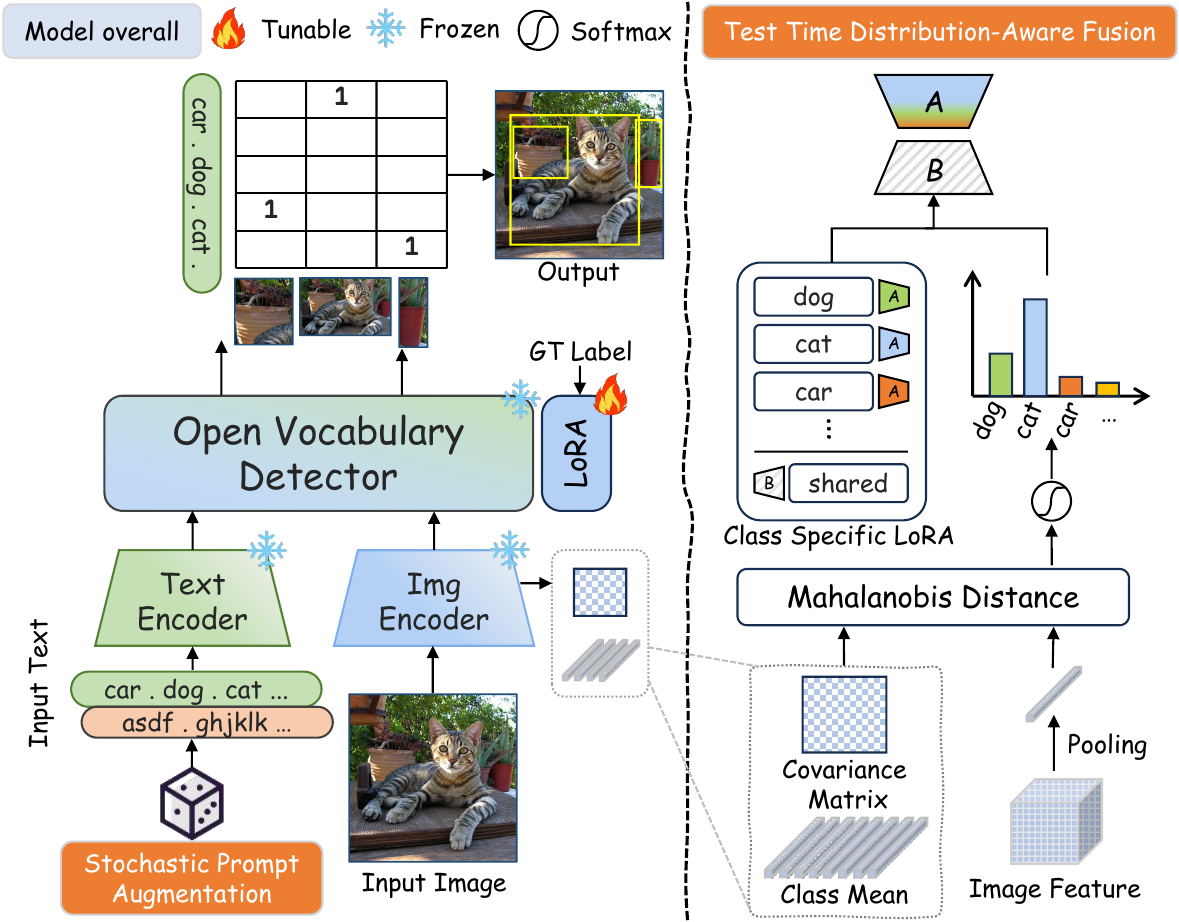}
    \caption{Overview of our proposed framework. (Left) The training architecture, incorporating Stochastic Prompt Augmentation (SPA) to simulate cross-task semantic interference during optimization. (Right) The Test-Time Distribution-Aware Fusion (TTDF) module, utilized exclusively during inference to dynamically weight and fuse class-specific LoRA experts.}
    \label{fig:framework}
\end{figure}

\noindent \textbf{Overview.}
\Cref{fig:framework} illustrates the overall architecture of our method, built upon the pre-trained open-vocabulary detector Grounding DINO~\cite{liu2024grounding}. To address the task-agnostic incremental learning challenge, our framework is decoupled into two distinct phases. During the training phase (\cref{fig:framework}, left), we maintain class-specific LoRA experts and a shared general knowledge module. To proactively align the training distribution with the complex inference scenario, we introduce a Stochastic Prompt Augmentation strategy that constructs cross-task and noisy text prompts dynamically during training. During the task-agnostic inference phase (\cref{fig:framework}, right), we propose a Test-Time Distribution-Aware Fusion mechanism. Instead of relying on uniform static averaging, this mechanism dynamically weights and aggregates the uncoordinated LoRA experts based on the probabilistic correspondence between the input visual features and the class mean, thereby mitigating parameter conflicts.

\subsection{Stochastic Prompt Augmentation}

\noindent \textbf{Revisiting Modular Adaptation.} 
To enable incremental adaptation without catastrophic forgetting, we build upon the modular framework of DitHub~\cite{cappellino2025dithub}, which conceptualizes incremental learning as the progressive expansion of a Low-Rank Adaptation (LoRA) library. Rather than employing standard LoRA updates, this framework decouples the adaptation into a shared projection matrix $\mathbf{B}$ and a set of class-specific experts $\{\mathbf{A}_c\}$. For a given class $c$, the adapted weight $\mathbf{W}$ is formulated as:
\begin{equation}
    \mathbf{W} = \mathbf{W}_0 + \mathbf{B} \mathbf{A}_c,
\end{equation}
where $\mathbf{W}_0$ denotes the frozen pre-trained weights of the base model, and $\mathbf{W}$ represents the final adapted weights for the target layer. During training, only the expert $\mathbf{A}_c$ corresponding to the ground-truth class is activated and updated. This strict parameter isolation effectively circumvents catastrophic forgetting, while the shared matrix $\mathbf{B}$ is continuously optimized across all tasks to maintain generalizable representations.

However, as analyzed in \cref{sec:ta_challenge}, a key limitation arises during task-agnostic evaluation.
The model is optimized with isolated, single-task prompts, but evaluated with a unified cross-task prompt containing all learned classes.
To address this mismatch between training and inference, we propose \textbf{Stochastic Prompt Augmentation (SPA).}
By stochastically constructing augmented training prompts, SPA simulates the prompt composition encountered during task-agnostic inference, thereby improving robustness to cross-task prompt interference.
Specifically, SPA introduces two types of distractors:

\noindent \textbf{Historical Replay Tokens.} To alleviate the model's sensitivity to cross-task interference, we introduce a dynamically sampled number of classes from previous tasks. This forces the model to discriminate the current objects from previously learned concepts within the unified textual context. 

\noindent \textbf{Synthetic Noise Tokens.} To fortify the model against the semantic interference of unseen classes from future incremental tasks and unknown concepts inherent to open-vocabulary queries, we inject a variable quantity of randomly generated strings. These synthetic tokens act as surrogates for out-of-vocabulary distractors. Crucially, by randomizing not only the textual content but also the exact number of injected tokens, we force the model to flexibly adapt to heterogeneous prompt structures, ensuring robust feature alignment regardless of unknown textual interference.

Formally, let $\mathcal{C}_{cur}$ denote the class set of the current task. During each training iteration, the augmented prompt set $\mathcal{P}_{train}$ is constructed as:
\begin{equation}
    \mathcal{P}_{train} = \mathcal{C}_{cur} \cup \mathcal{C}'_{prev} \cup \mathcal{C}_{noise},
\end{equation}
where $\mathcal{C}'_{prev}$ and $\mathcal{C}_{noise}$ denote the stochastically sampled past classes and synthetic noise strings, respectively. These additional tokens are introduced for multimodal image–text fusion to enhance robustness to semantic interference and are therefore excluded from object matching and loss computation.

\subsection{Test-Time Distribution-Aware Fusion}

\noindent \textbf{Inference via Merging.} 
During inference, DitHub dynamically retrieves the corresponding LoRA modules from its library according to the classes specified in the input prompt. The final adaptation is obtained by merging these modules, directly through a naive arithmetic mean operation:
\begin{equation}
    \mathbf{W} = \mathbf{W}_0 + \mathbf{B}\left( \frac{1}{|\mathcal{P}|} \sum_{c \in \mathcal{P}} \mathbf{A}_c \right),
    \label{eq:dithub_merge}
\end{equation}
where $\mathcal{P}$ denotes the set of target classes present in the current inference prompt.

While the static parameter averaging employed by DitHub proves effective for isolated tasks—where intra-task experts share consistent optimization trajectories—indiscriminately merging LoRA modules across heterogeneous tasks inherently disrupts learned weights and triggers severe parameter conflicts. Motivated by this critical vulnerability, we propose a \textbf{Test-Time Distribution-Aware Fusion (TTDF)} mechanism. Rather than treating all class-specific experts equally, TTDF dynamically aggregates parameters by evaluating the likelihood that an input image aligns with the learned feature distribution of each expert. In particular, we maintain a probabilistic profile (comprising the mean and covariance) for each LoRA module, enabling us to precisely measure the visual correspondence and selectively activate the most relevant experts during inference.

\noindent \textbf{Online Distribution Modeling.} Formally, for each class-specific LoRA module, we track a class mean vector $\boldsymbol{\mu}_{c}$ (as the retrieval key) and a counter $t_{c}$. To characterize the global intra-class variance, we additionally estimate a shared scatter matrix $\mathbf{S}$. All these statistics are computed in a streaming manner during training via an online update rule.

Given an input image $\mathbf{x}$ processed by the image encoder $f(\cdot)$ with ground-truth classes $\mathcal{C}_{gt} = \{c_1, \dots, c_m\}$, for each class $c \in \mathcal{C}_{gt}$, we update the statistics as follows:
\begin{equation}\boldsymbol{\mu}_{c}^{(t)} = \boldsymbol{\mu}_{c}^{(t-1)} + \frac{1}{t_{c}} (f(\mathbf{x}) - \boldsymbol{\mu}_{c}^{(t-1)}),
\end{equation}
\begin{equation}
\mathbf{S}^{(t)} = \mathbf{S}^{(t-1)} + (f(\mathbf{x}) - \boldsymbol{\mu}_{c}^{(t-1)})(f(\mathbf{x}) - \boldsymbol{\mu}_{c}^{(t)})^\top.
\end{equation}
The shared covariance matrix $\boldsymbol{\Sigma}$ is then derived by normalizing the scatter matrix: 

\begin{equation}
\boldsymbol{\Sigma} = \frac{1}{\sum t_c - 1} \mathbf{S}.
\end{equation}

\noindent \textbf{Mahalanobis Distance-based Reweighting.}
During task-agnostic inference, for a test image $\mathbf{x}$, we compute the relevance score of each LoRA module using the Mahalanobis distance~\cite{lee2018simple,goswami2023fecam}. Unlike standard Euclidean or cosine metrics that treat all feature dimensions isotropically, the Mahalanobis distance leverages the estimated covariance to capture the true anisotropic distribution of the feature space. This effectively downweights noisy dimensions and accounts for feature correlations:

\begin{equation}
d_c(\mathbf{x}) = (f(\mathbf{x}) - \boldsymbol{\mu}_c) \boldsymbol{\Sigma}^{-1} (f(\mathbf{x}) - \boldsymbol{\mu}_c)^\top.
\end{equation}
Since a smaller distance indicates a higher likelihood, we convert these distances into normalized importance weights using the Softmax function with a temperature $\boldsymbol{\tau}$:
\begin{equation}
w_c(\mathbf{x}) = \frac{e^{-d_c(\mathbf{x})/\boldsymbol{\tau}}}{\sum_{j \in \mathcal{P}_{TA}} e^{-d_j(\mathbf{x})/\boldsymbol{\tau}}},
\end{equation}
where $\mathcal{P}_{TA}$ denotes the set of all learned classes. The temperature $\boldsymbol{\tau}$ explicitly controls the sharpness of this weight distribution.

Finally, the adapted weights are computed by dynamically fusing the class-specific $\mathbf{A}$ matrices based on the input-specific weights $w_c(\mathbf{x})$, while sharing the $\mathbf{B}$ matrix:
\begin{equation}
\mathbf{W} =\mathbf{W}_0 + \mathbf{B}\left( \sum_{c \in \mathcal{P}_{TA}} w_c(\mathbf{x}) \mathbf{A}_c \right).
\end{equation}
This mechanism ensures that the final adaptation is dominated heavily by the most relevant experts while suppressing the noisy contributions of unrelated classes, effectively resolving the parameter conflict in the TA-IVLOD setting.

\section{Experiments}
\subsection{Experimental Setup}

\noindent \textbf{Evaluation settings.}
To comprehensively evaluate our proposed framework, we conduct experiments under two distinct protocols. First, to align with realistic open-world scenarios, we evaluate under our proposed TA-IVLOD setting. In this challenging setup, the model must evaluate all test samples using a unified, cross-task prompt encompassing all encountered classes, without any task identity priors. Second, we use the standard IVLOD setting, where the model relies on a task oracle during inference to construct sparse, task-specific prompts.

Following previous works~\cite{deng2024zero,cappellino2025dithub}, we conduct experiments on the Object Detection in the Wild (ODinW~\cite{li2022grounded}) benchmark and the MS COCO~\cite{lin2014microsoft} dataset. ODinW is a challenging benchmark designed to test model performance under significant domain shifts in real-world scenarios. We primarily employ \textbf{ODinW-13}, which comprises 13 diverse sub-datasets spanning from traditional natural images (e.g., Pascal VOC~\cite{everingham2010pascal}) to highly specialized domains (e.g., Thermal imagery, Aerial Maritime Drones, and Aquariums). Specifically, each sub-dataset is formulated as an individual task within the incremental learning process, where the model sequentially learns across these 13 tasks.

\begin{table*}[t]
\centering
\caption{Quantitative comparison of mAP performance on ODinW-13 under the TA-IVLOD settings. Avg represents the average mAP across all 13 sequentially learned downstream tasks, while zero-shot generalization performance is reported in the $\text{Z}_\text{coco}$ column. \textbf{G-Dino} represents the zero-shot performance of Grounding DINO. \textbf{Bold} and \underline{underline} indicate the best and second-best results, respectively.}
\label{tab:main_ta}
\resizebox{\textwidth}{!}{
\begin{tabular}{c|cc|ccccccccccccc}
\toprule
 Method & $\text{Z}_\text{coco}$ & Avg  & Ae   & Aq   & Co   & Eg   & Mu   & Pa   & Pv   & Pi   & Po   & Ra   & Sh   & Th   & Ve   \\
\midrule
  G-Dino & 48.4  & 24.0 & 15.9 & 11.5 & 16.3 & 28.1 & 21.0 & 5.2  & 46.7 & 36.4 & 2.4  & 0.0  & 20.4 & 55.3 & 53.0 \\
  \midrule
                            TFA    & 42.7  & 22.0 & 15.0 & 10.1 & 16.9 & 25.9 & 22.2 & 4.1  & 43.6 & 29.5 & 1.6  & 0.0  & 16.2 & 50.7 & 50.4 \\
                            iDETR  &37.4 & 34.0 & \underline{30.1} & 30.7 & 44.1 & \underline{40.6} & 31.6 & 7.0  & 58.9 & 37.0 & 10.8 & \underline{0.2}  & 30.3 & 63.5 & 57.4 \\
                            ZiRa   & 45.1  & \underline{35.0} & 28.5 & \underline{31.4} & \underline{46.5} & 34.6 & 35.8 & 4.6  & 59.0 & 37.6 & \underline{15.5} & 0.1  & \underline{32.7} & 65.7 & \underline{62.4} \\
                            DitHub & \underline{46.8}  & 32.5 & 23.8 & 16.8 & 38.2 & 30.6 & \underline{37.5} & \underline{8.5}  & \underline{63.2} & \underline{40.2} & 3.8  & 0.0  & 30.2 & \underline{67.9} & 62.2 \\
                            ours   & \textbf{47.2}  & \textbf{64.5} & \textbf{43.4} & \textbf{50.3} & \textbf{82.0} & \textbf{70.6} & \textbf{51.0} & \textbf{89.0} & \textbf{69.3} & \textbf{68.6} & \textbf{53.3} & \textbf{61.4} & \textbf{56.2} & \textbf{71.4} & \textbf{71.3} \\
\bottomrule
\end{tabular}
}
\end{table*}

For quantitative evaluation, we adopt mean Average Precision (mAP), measured after the model finishes sequential training on all sub-datasets. We report two primary metrics: \textbf{Avg} and \textbf{$\text{Z}_\text{coco}$}.
Specifically, Avg represents the average mAP over all downstream tasks in ODinW-13. Under the TA-IVLOD protocol, following the established evaluation paradigm of traditional incremental object detection~\cite{zhang2025few,liu2023augmented,kim2024vlm,zhang2024learning,luo2025gradient}, the individual test sets of all 13 sub-datasets are merged into a unified test set. On this merged test set, an AP is computed for each unique class. For classes shared across multiple tasks, a global AP is computed and applied consistently to the respective task-level mAP calculations.

In addition, we report $\text{Z}_\text{coco}$, the zero-shot mAP on the MS COCO dataset. Since COCO covers a broad and diverse visual domain, this metric evaluates whether the open-vocabulary detector preserves its inherent zero-shot generalization ability after sequential fine-tuning.

We compare our framework against several state-of-the-art methods. Our primary baselines are ZiRa~\cite{deng2024zero} and DitHub~\cite{cappellino2025dithub}, which are specifically tailored for IVLOD. Additionally, we benchmark against methods designed for Incremental Learning (TFA~\cite{wang2020frustratingly}, iDETR~\cite{dong2023incremental}).

\noindent \textbf{Implementation details.} Unless otherwise specified, we employ Grounding DINO with a Swin-Tiny backbone as our default Vision-Language Object Detection Model. This base model is pre-trained on Objects365~\cite{shao2019objects365}, GoldG~\cite{li2022grounded}, and Cap4M~\cite{li2022grounded}. To enable incremental learning, we build our pipeline upon the DitHub framework. During sequential training, we follow its LoRA-based optimization strategy and corresponding hyperparameter settings to cultivate class-specific experts for all newly encountered classes.

\subsection{Comparison Results}

\begin{table*}[t]
\centering
\caption{Quantitative comparison of mAP performance on ODinW-13 under the standard IVLOD setting. \textbf{G-Dino} represents the zero-shot performance of Grounding DINO. \textbf{Bold} and \underline{underline} indicate the best and second-best results, respectively.}
\label{tab:main_ivlod}
\resizebox{\textwidth}{!}{
\begin{tabular}{c|cc|ccccccccccccc}
\toprule
Method & $\text{Z}_\text{coco}$ & Avg  & Ae   & Aq   & Co   & Eg   & Mu   & Pa   & Pv   & Pi   & Po   & Ra   & Sh   & Th   & Ve   \\
\midrule
 G-Dino & 48.4  & 52.9 & 19.2 & 20.4 & 80.0 & 59.7 & 33.6 & 81.1 & 56.4 & 67.6 & 27.2 & 68.4 & 38.7 & 74.1 & 61.3 \\
 \midrule
                            TFA    & 42.7  & 50.1 & 18.0 & 19.5 & 76.7 & 52.3 & 34.5 & 79.7 & 52.6 & 64.3 & 21.4 & 67.5 & 32.2 & 74.3 & 58.4 \\
                            iDETR  &37.4 & 58.4 & 35.1 & 40.3 & 79.3 & 59.7 & 47.6 & 82.6 & 62.6 & 66.0 & 39.3 & 67.4 & 46.9 & 75.1 & 57.0      \\
                            ZiRa   & 45.1  & 60.6 & 34.6 & 43.1 & 82.0 & 63.9 & 48.7 & 82.4 & 64.5 & 69.3 & 46.4 & 68.9 & 44.8 & 77.3 & 62.2 \\
                            DitHub & \underline{46.8}  & \underline{66.3} & \underline{35.9} & \underline{51.9} & \underline{83.2} & \textbf{71.6} & \textbf{56.4} & \underline{87.4} & \underline{70.0} & \underline{70.6} & \underline{54.7} & \underline{73.4} & \underline{55.2} & \underline{82.9} & \underline{69.2} \\
                            ours   & \textbf{47.2}  & \textbf{67.5} & \textbf{36.9} & \textbf{54.7} & \textbf{85.1} & \underline{71.3} & \underline{53.6} & \textbf{89.2} & \textbf{71.0} & \textbf{71.8} & \textbf{55.1} & \textbf{74.8} & \textbf{58.7} & \textbf{84.6} & \textbf{71.0} \\
\bottomrule
\end{tabular}
}
\end{table*}

\begin{table*}[t]
\centering
\caption{{Quantitative results of few-shot (1, 5, and 10 shots) incremental object detection on ODinW-13. $\text{Avg}_\text{I}$ and $\text{Avg}_\text{TA}$ denote the average mAP across all 13 downstream tasks under IVLOD and TA-IVLOD settings, respectively.}}
\label{tab:few_shot}

\resizebox{0.8\textwidth}{!}{

\begin{tabular}{c|ccc|ccc|ccc}
\toprule
\multirow{2}{*}{Method} & \multicolumn{3}{c|}{1 shot}                        & \multicolumn{3}{c|}{5 shot}                        & \multicolumn{3}{c}{10 shot}                       \\
                        & $\text{Z}_\text{coco}$ & $\text{Avg}_\text{I}$ & $\text{Avg}_\text{TA}$ & $\text{Z}_\text{coco}$ & $\text{Avg}_\text{I}$ & $\text{Avg}_\text{TA}$ & $\text{Z}_\text{coco}$ & $\text{Avg}_\text{I}$ & $\text{Avg}_\text{TA}$ \\
\midrule
TFA                     & 45.1                      & 50.5  & 23.0     & 44.4                      & 51.0  & 22.9     & 44.7                      & 50.9  & 22.9     \\
iDETR                   & 45.6                      & 52.7  & 27.7     & 41.8                      & 55.5  & 29.3     & 42.7                      & 56.8  & 31.7     \\
ZiRa                    & 46.5                      & 54.1  & 27.1     & 45.2                      & 56.9  & 32.1     & 45.7                      & 57.1  & 30.1     \\
DitHub                  & 45.9                      & 54.9  & 26.2     & 44.1                      & 59.8  & 28.3     & 45.2                      & 60.5  & 30.3     \\
ours                    & 46.7                      & 56.2  & 33.7     & 44.1                      & 61.2  & 45.7     & 45.5                      & 61.0  & 53.9\\
\bottomrule

\end{tabular}
}
\end{table*}
All reported quantitative results are averaged over three independent runs with different random seeds to ensure statistical reliability. Prior baselines, including TFA, iDETR, ZiRa, and DitHub, are re-implemented under our unified evaluation protocol. The discrepancies between our reproduced results and those reported in \cite{deng2024zero} are discussed in detail in the supplementary material.

\noindent \textbf{Performance in the TA-IVLOD Setting.} As shown in \cref{tab:main_ta}, our proposed method demonstrates robust performance against cross-task semantic interference, yielding a state-of-the-art average mAP of 64.5 in the challenging TA-IVLOD setting. This corresponds to a +32.0 absolute improvement over DitHub. This substantial gain indicates that the synergistic design of SPA and TTDF effectively alleviates parameter conflicts and cross-task semantic interference.

In contrast, prior methods experience significant performance degradation under this unconstrained setting, often approaching near-zero performance on certain tasks (\eg, Raccoon (Ra) and Packages (Pa)). This behavior is largely attributed to the synthetic long-tailed distribution introduced by aggregating test sets from distinct tasks. Without explicit task boundaries, models optimized solely with single-task prompts struggle to suppress irrelevant cross-task distractors. Consequently, for small-scale tasks with limited positive instances, increased false-positive rates substantially degrade the precision metric.

\noindent \textbf{Performance in the Standard IVLOD Setting.} Beyond the improvements observed in the task-agnostic scenario, our approach maintains strong detection performance under the standard IVLOD setting. As detailed in \cref{tab:main_ivlod}, our method achieves an average mAP of 67.5, outperforming DitHub by +1.2 mAP. This consistent improvement indicates that our TTDF module effectively mitigates parameter conflicts among class-specific experts even when oracle task identities are provided during inference.

\noindent \textbf{Robustness Across Few-Shot Regimes.} To further assess generalization, we evaluate its performance under strictly limited data conditions. As shown in \cref{tab:few_shot}, our approach consistently outperforms existing baselines across 1-shot, 5-shot, and 10-shot configurations. Notably, the performance advantage is maintained in both the standard IVLOD ($\text{Avg}_\text{I}$) and the TA-IVLOD ($\text{Avg}_\text{TA}$) settings, underscoring the general efficacy and stability of our proposed modules across varying data scales.

\subsection{Model Analysis and Ablation Studies}

\noindent \textbf{Module Ablation.}
We investigate the individual and combined efficacy of our core modules on the ODinW-13 benchmark, as detailed in \cref{tab:module_ablation}. Compared to the baseline (DitHub), which suffers significant performance degradation in the task-agnostic setting, incorporating TTDF yields a notable improvement of +12.7 mAP, demonstrating its effectiveness in mitigating dynamic parameter conflicts. Furthermore, integrating Stochastic Prompt Augmentation (SPA) on top of TTDF brings an additional gain of +18.4 mAP. This demonstrates that synergizing train-time textual augmentation with test-time dynamic routing facilitates more discriminative target identification under complex semantic interference, ultimately improving the overall performance to 64.5 mAP.

\begin{table}[!t]
\begin{minipage}[t]{0.51\linewidth}
    \centering
    \caption{{Ablation study of TTDF and SPA  on ODinW-13.}}
    \label{tab:module_ablation}
\resizebox{0.77\textwidth}{!}{
\begin{tabular}{c|cc|cc}
\toprule
 & TTDF & SPA & $\text{Z}_\text{coco}$ & $\text{Avg}_\text{TA}$    \\
\midrule
baseline &              &              & 46.8 & 32.5 \\
\midrule
\multirow{2}{*}{ours}     &\checkmark    &              & 47.4 & 45.9 \\
         &\checkmark    & \checkmark   & 47.2 & 64.5 \\
\bottomrule

\end{tabular}
}
\end{minipage}
  \hfill
\begin{minipage}[t]{0.45\linewidth}
    \centering
    \caption{Ablation study of different prompt augmentation strategies.}
    \label{tab:prompt_ablation}
\resizebox{\textwidth}{!}{
\begin{tabular}{c|cc}
\toprule
                           & $\text{Z}_\text{coco}$ & $\text{Avg}_\text{TA}$ \\
\midrule
Learned classes            & 47.0 & 57.2    \\
Noise tokens               & 47.4 & 60.3    \\
COCO                       & 46.0 & 58.2    \\
Learned class + Noise tokens & 47.2  & 64.5    \\
\bottomrule
\end{tabular}
}
\end{minipage}

\end{table}

\noindent \textbf{Prompt Augmentation Strategies.} 
\Cref{tab:prompt_ablation} compares four augmentation strategies utilizing different distractor types: historical learned classes, synthetic noise tokens, external COCO classes, and our combined approach. Specifically, while learned classes provide a basic level of robustness, the gains are constrained by the limited diversity of previously seen concepts. Although introducing COCO classes enhances task-agnostic resistance in downstream tasks, it leads to a performance decline in the zero-shot detection of those specific classes due to the semantic confusion induced during training. Employing any real-world classes as distractors inherently impairs the model's ability to recognize those specific classes, as it is forced to treat valid semantic concepts as negative interference. In contrast, synthetic noise tokens offer a unique advantage by providing textual complexity without concrete semantic meaning, thus preserving the zero-shot capabilities of real-world classes. By combining historical classes with synthetic noise, our method successfully emulates the complex, varied-length prompts encountered during actual inference. This synergy between realistic distractors and neutral noise yields the best performance across all metrics.

\noindent \textbf{Mahalanobis Distance.}
\Cref{tab:dis_ablation} compares the effectiveness of different distance metrics within our TTDF module. As illustrated, employing Mahalanobis distance yields substantial improvements over the standard Euclidean distance. The effectiveness of Mahalanobis distance stems from its ability to account for the underlying data distribution and feature correlations. Euclidean distance assumes an isotropic spherical distribution, which often fails to capture true semantic proximity in high-dimensional latent spaces characterized by varying scales and feature inter-dependencies. By contrast, Mahalanobis distance effectively normalizes these variances and incorporates the covariance structure of the feature clusters. This allows the TTDF module to more accurately identify and retrieve the most relevant class-specific experts during inference, thereby substantially mitigating parameter conflicts and enhancing the overall discriminative power of the model.

\begin{table}[!t]
\begin{minipage}[t]{0.42\linewidth}
    \centering
    \caption{{Comparison of different distance metrics for TTDF.}}
    \label{tab:dis_ablation}
\resizebox{0.9\textwidth}{!}{
\begin{tabular}{cccc}
\toprule
     & $\text{Z}_\text{coco}$ & $\text{Avg}_\text{TA}$  \\
\midrule
Euclidean Distance & 45.9                      & 51.3                              \\
Mahalanobis Distance & 47.2                      & 64.5                              \\
\bottomrule
\end{tabular}
}
\end{minipage}
  \hfill
\begin{minipage}[t]{0.53\linewidth}
    \centering
    \caption{Comparison with task-agnostic baselines under IOD and TA-IVLOD settings.}
    \label{tab:iod}
\resizebox{0.8\textwidth}{!}{
\begin{tabular}{ccccc}
\toprule
 & G-Dino & GCD  & DitHub & ours \\
\midrule
IOD (40-40)                 & 48.4   & 55.2 & 53.2   & 54.5 \\
TA-IVLOD             & 24.0  & 29.0   & 32.5   & 64.5 \\
\bottomrule
\end{tabular}

}
\end{minipage}

\end{table}%

\noindent \textbf{Generalization to Incremental Object Detection.} Conventional Incremental Object Detection (IOD) aims to incrementally learn
new object categories while retaining previously learned ones, and has been
widely studied in class-incremental detection scenarios~\cite{dong2023incremental,zhang2026yolo,
kim2024vlm,wang2025gcd,zhang2024learning}. Unlike our TA-IVLOD setting,
which additionally involves open-vocabulary detectors, heterogeneous downstream
domains, and unified cross-task prompts without task-specific priors, conventional
IOD is typically evaluated within a fixed dataset and a closed category space.
To further validate the generalizability of our method beyond the IVLOD benchmark, we additionally conduct experiments under the conventional Incremental Object Detection (IOD) setting on MS COCO. Specifically, the 80 COCO categories are split into two sequential learning stages, with 40 classes introduced at each stage. In this setting, images may naturally contain objects from different incremental stages, requiring the detector to handle mixed-category scenes after sequential learning. As shown in \cref{tab:iod}, our method remains competitive with the recent task-agnostic IOD method GCD\cite{wang2025gcd} under the conventional IOD protocol, while outperforming other baselines. More importantly, our method exhibits a clear advantage over GCD under the more challenging TA-IVLOD setting, where diverse task domains are evaluated jointly without task-specific priors. This demonstrates that our framework not only generalizes well to conventional IOD scenarios, but also remains effective in more heterogeneous task-agnostic environments.

\section{Conclusion}
In this paper, we identify a critical limitation in current IVLOD research: the over-reliance on oracle task identities, which restricts practical open-world deployment. To address this, we formalize TA-IVLOD, a more realistic setting where task boundaries are unknown during inference. We demonstrate that existing methods suffer from severe performance degradation in this setting due to prompt-induced semantic interference.

To overcome these challenges, we propose a modular LoRA-based framework featuring a Test-Time Distribution-Aware Fusion mechanism and a Stochastic Prompt Augmentation strategy. These components effectively mitigate parameter conflicts and enhance cross-modal robustness against cross-task, unified prompts. Extensive evaluations on the ODinW-13 benchmark prove that our method not only excels in standard IVLOD but also yields substantial improvements in the challenging TA-IVLOD setting, while effectively preserving pre-trained zero-shot capabilities. 

Despite its effectiveness, the dynamic nature of our test-time fusion introduces a slight computational overhead during inference. Additionally, while our current fusion mechanism relies on image-level features for dynamic weighting, exploring instance-level features presents a promising avenue for future work.
\section*{Acknowledgements}
This work is funded by NSFC (NO. 62206135, 62225604),
Young Elite Scientists Sponsorship Program by CAST (2023QNRC001), the Fundamental Research Funds for the Central Universities (Nankai University, 070-63263251), Guangdong Basic and Applied Basic Research Foundation (No. 2026A1515011147), Shenzhen Science and Technology Program (JCYJ20240813114237048).

\bibliographystyle{splncs04}
\bibliography{main}

\clearpage
\appendix
\section{Detailed Benchmark Protocols}

To evaluate our proposed framework in both the standard IVLOD and the TA-IVLOD settings, we follow the evaluation protocol established by Deng et al. \cite{deng2024zero}. Specifically, we conduct our experiments on the ODinW-13 benchmark and the MS COCO dataset.

The ODinW-13 benchmark is designed to assess model robustness and adaptability across significant domain shifts and diverse real-world scenarios. It aggregates data from thirteen distinct downstream sources, which are formulated as sequential tasks in our incremental learning setup. The constituent datasets and their corresponding abbreviations are as follows: Aerial Maritime Drone (Ae), Aquarium (Aq), Cottontail Rabbit (Co), EgoHands (Eg), Mushrooms (Mu), Packages (Pa), Pascal VOC (Pv), Pistols (Pi), Pothole (Po), Raccoon (Ra), Shellfish (Sh), Thermal Dogs and People (Th), and Vehicles (Ve).

This benchmark encompasses a wide spectrum of visual challenges, ranging from standard natural scenes to highly specialized domains such as thermal and aerial imagery. The high diversity of these 13 datasets provides a rigorous testbed for evaluating the model's capacity to handle cross-task semantic interference under the task-agnostic setting.

\subsection{Correction of ODinW-13 Annotation IDs and Evaluation Discrepancy}
\label{sec:annotation_correction}

In this section, we detail an important correction applied to the ODinW-13 test sets, which accounts for the performance discrepancies between our reproduced baseline results and those originally reported by Deng et al. \cite{deng2024zero}. 

Specifically, we identified a formatting incompatibility regarding the ground-truth annotation IDs in the ODinW-13 benchmark. With the exception of the Pistols (Pi) dataset, the remaining 12 datasets contain bounding boxes with an \texttt{annotation\_id} of 0. Because current IVLOD methods rely on the standard MS COCO evaluation API—which assumes 1-based indexing for annotations—any ground-truth instance with an ID of 0 is inherently excluded during evaluation. Consequently, if a model correctly localizes these objects, the predictions are counted as False Positives (FPs) due to the missing ground-truth references, leading to an underestimation of the true performance.

To address this, we corrected the annotation files for the affected datasets by reassigning all instances where \texttt{annotation\_id} = 0 to \texttt{max\_annotation\_id} + 1. Following this adjustment, we re-evaluated the baseline methods and observed that their actual performance is notably higher than previously reported. All quantitative results presented in our main paper reflect this corrected evaluation protocol.

As shown in \cref{tab:dataset}, correcting the annotation IDs improves the measured performance across all methods. The impact is particularly pronounced on datasets with smaller test sets, such as Packages (Pa), where the omission of 0-ID instances exerts a magnified penalty on the mAP. For example, the score for DitHub increases substantially from 64.9 to 87.4. As expected, performance on the properly indexed Pistols (Pi) dataset remains unchanged. Ultimately, this correction elevates the overall average mAP ($\text{Avg}_\text{I}$) by 3.5 to 4.1 points across the board, providing a more accurate reflection of model capabilities.

\begin{table*}[t]
\centering
\caption{Quantitative comparison of mAP performance on the ODinW-13 benchmark, demonstrating the evaluation discrepancy caused by formatting inconsistencies. \textbf{Original} and \textbf{Rectified} denote the results before and after the test set annotation correction, respectively. $\text{Avg}_\text{I}$ represents the average mAP across all 13 downstream tasks under the standard IVLOD setting.}
\label{tab:dataset}
\resizebox{\textwidth}{!}{
\begin{tabular}{c|c|c|ccccccccccccc}
\toprule
Method                  & Annotations & $\text{Avg}_\text{I}$  & Ae   & Aq   & Co   & Eg   & Mu   & Pa   & Pv   & Pi   & Po   & Ra   & Sh   & Th   & Ve   \\
\midrule
\multirow{2}{*}{ZiRa}   & Original  & 57.1 & 34.2 & 40.9 & 69.1 & 63.9 & 48.6 & 62.2 & 64.4 & 69.3 & 45.8 & 64.9 & 44.1 & 74.5 & 61   \\
                        & Rectified   & 60.6 & 34.6 & 43.1 & 82.0 & 63.9 & 48.7 & 82.4 & 64.5 & 69.3 & 46.4 & 68.9 & 44.8 & 77.3 & 62.2 \\
\midrule
\multirow{2}{*}{DitHub} & Original  & 62.4 & 35.5 & 49.4 & 69.8 & 71.6 & 56.2 & 64.9 & 69.9 & 70.6 & 53.7 & 69.5 & 53.9 & 78.4 & 67.6 \\
                        & Rectified   & 66.3 & 35.9 & 51.9 & 83.2 & 71.6 & 56.4 & 87.4 & 70.0 & 70.6 & 54.7 & 73.4 & 55.2 & 82.9 & 69.2 \\
\midrule
\multirow{2}{*}{ours}   & Original  & 63.4 & 36.6 & 52.2 & 72.5 & 71.3 & 53.5 & 63.2 & 70.9 & 71.8 & 54.2 & 70.6 & 57.7 & 80.1 & 69.4 \\
                        & Rectified   & 67.5 & 36.9 & 54.7 & 85.1 & 71.3 & 53.6 & 89.2 & 71.0 & 71.8 & 55.1 & 74.8 & 58.7 & 84.6 & 71.0 \\
\bottomrule
\end{tabular}
}
\end{table*}

\subsection{Evaluation on ODinW Subset for Domain Overlap}
\label{sec:odinw_subset}

In the TA-IVLOD setting, merging the test sets of all tasks can introduce unannotated cross-task objects. Consequently, valid detections of these objects are incorrectly penalized as False Positives (FPs), which skews the evaluation metrics. To mitigate this evaluation bias, we construct ODinW-S, a domain-exclusive subset comprising seven datasets from the ODinW-13 benchmark: Aerial Maritime Drone (Ae), Aquarium (Aq), Cottontail Rabbit (Co), Mushrooms (Mu), Packages (Pa), Raccoon (Ra), and Thermal Dogs and People (Th). Characterized by non-overlapping categories and distinct visual domains, these datasets effectively circumvent the issue of missing cross-task annotations.

\begin{table*}[t]
\centering
\caption{Quantitative comparison of mAP performance on ODinW-S under both standard IVLOD and TA-IVLOD settings. G-Dino represents the zero-shot performance of Grounding DINO.}
\label{tab:oindw_s}
\resizebox{0.8\textwidth}{!}{
\begin{tabular}{c|c|c|c|ccccccc}
\toprule
Setting                   & Method & $\text{Z}_\text{coco}$ & Avg  & Ae   & Aq   & Co   & Mu   & Pa   & Ra   & Th   \\
\midrule
\multirow{6}{*}{IVLOD}    & G-Dino  & 48.4  & 53.8 & 19.2 & 20.4 & 80.0 & 33.6 & 81.1 & 68.4 & 74.1 \\
                          & TFA    & 46.2  & 49.9 & 19.1 & 22.5 & 80.7 & 34.0 & 83.7 & 65.0 & 75.6 \\
                          & iDETR  & 41.8  & 60.6 & 39.2 & 45.2 & 81.8 & 52.2 & 80.1 & 72.2 & 76.5 \\
                          & ZiRa   & 44.6  & 64.2 & 36.5 & 42.9 & 83.2 & 50.0 & 85.5 & 71.1 & 80.1 \\
                          & DitHub & 44.7  & 64.1 & 39.6 & 52.3 & 84.3 & 56.5 & 88.0 & 77.8 & 85.2 \\
                          & ours   & 44.8  & 68.8 & 37.4 & 55.7 & 83.3 & 51.6 & 90.0 & 77.2 & 86.2 \\
\midrule
\multirow{6}{*}{TA-IVLOD} & G-Dino  & 48.4  & 25.3 & 13.6 & 15.0 & 53.8 & 34.4 & 52.5 & 0.0  & 7.8  \\
                          & TFA   & 46.2  & 24.9 & 13.4 & 14.3 & 50.9 & 33.7 & 52.9 & 0.0  & 8.8  \\
                          & iDETR  & 41.8  & 39.6 & 31.9 & 39.1 & 25.4 & 45.5 & 67.0 & 36.6 & 31.9 \\
                          & ZiRa   & 44.6  & 37.9 & 31.6 & 38.1 & 50.6 & 49.7 & 51.2 & 12.4 & 31.7 \\
                          & DitHub & 44.7  & 29.9 & 22.9 & 30.4 & 49.7 & 41.8 & 44.5 & 0.0  & 20.2 \\
                          & ours   & 44.8  & 62.0 & 36.2 & 52.5 & 78.7 & 51.2 & 90.0 & 53.2 & 71.9 \\
\bottomrule
\end{tabular}
}
\end{table*}

As shown in \cref{tab:oindw_s}, even when domain overlap is strictly controlled, existing baselines still suffer from substantial performance degradation under the TA-IVLOD setting. In contrast, our approach maintains a robust average mAP of 62.0. This confirms that the proposed method effectively mitigates prompt-induced semantic interference, demonstrating a genuine robustness that is independent of dataset annotation artifacts.

\section{In-depth Experimental Analysis}

\subsection{Impact of Prompt Interference}
\label{sec:prompt_interference}

To explicitly isolate and quantify the semantic interference induced by cross-task prompts, we design a targeted evaluation protocol. In this setup, we evaluate the models on the independent test sets of each task, identical to the standard IVLOD setting. However, during inference, we provide the models with a unified prompt encompassing all categories from all sequentially learned tasks. We then filter the final predictions during the metric computation, strictly retaining only the subset of categories relevant to the specific task being evaluated.

\begin{table*}[h]
\centering
\caption{Quantitative evaluation of prompt-induced semantic interference on ODinW-13. Models are tested on task-specific sets but provided with a unified prompt encompassing all categories.  \textbf{G-Dino} denotes the zero-shot performance of Grounding DINO.}
\label{tab:separation}
\resizebox{0.9\textwidth}{!}{
\begin{tabular}{c|c|ccccccccccccc}
\toprule
Method & Avg  & Ae   & Aq   & Co   & Eg   & Mu   & Pa   & Pv   & Pi   & Po   & Ra   & Sh   & Th   & Ve   \\
\midrule
G-Dino & 41.6 & 17.7 & 17.5 & 77.2 & 58.1 & 36.9 & 82.1 & 49.6 & 47.2 & 18.5 & 10.1 & 21.2 & 46.5 & 57.7 \\
\midrule
TFA    & 38.1 & 17.9 & 15.2 & 76.6 & 52.0 & 37.6 & 73.2 & 46.2 & 41.3 & 13.6 & 8.2  & 17.2 & 43.0 & 53.4 \\
iDETR  & 49.7 & 31.8 & 38.9 & 77.3 & 57.5 & 41.9 & 85.5 & 61.6 & 43.5 & 26.0 & 30.2 & 32.7 & 62.5 & 56.4 \\
ZiRa   & 51.8 & 32.0 & 42.1 & 80.5 & 62.7 & 45.3 & 86.0 & 62.7 & 44.1 & 40.1 & 19.8 & 34.5 & 61.1 & 62.2 \\
DitHub & 47.5 & 21.5 & 26.1 & 78.9 & 56.3 & 42.0 & 84.0 & 65.6 & 47.3 & 20.6 & 18.2 & 30.6 & 65.0 & 60.8 \\
ours   & 66.0 & 36.4 & 54.0 & 84.9 & 71.3 & 51.1 & 89.0 & 70.9 & 69.0 & 54.1 & 70.0 & 57.3 & 79.0 & 71.2 \\
\bottomrule
\end{tabular}
}
\end{table*}

The results in \cref{tab:separation} effectively quantify the impact of prompt-induced semantic interference. By filtering out cross-task predictions during evaluation, we eliminate the influence of potential false positives (FPs) discussed in \cref{sec:odinw_subset}. The significant performance degradation observed in baseline methods, yielding average mAPs of 51.8 for ZiRa and 47.5 for DitHub, reveals that prompt interference inherently hinders the detection of target categories rather than merely introducing more FPs. In contrast, our proposed method effectively suppresses this semantic noise, maintaining a robust average mAP of 66.0. This demonstrates that our approach preserves the discriminative power of the model even when exposed to irrelevant cross-task categories.

\subsection{Sensitivity to Temperature $\boldsymbol{\tau}$}
\label{sec:hyperparameter}
\begin{figure}[t]
    \centering
    \includegraphics[width=0.6\textwidth]{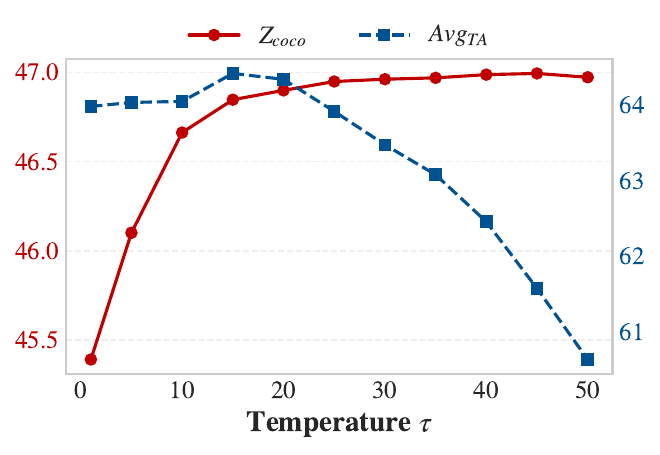}
    \caption{Sensitivity analysis of the temperature parameter $\boldsymbol{\tau}$ in TTDF.}
    \label{fig:temperature}
\end{figure}

In the Test-Time Distribution-Aware Fusion (TTDF) module, the temperature parameter $\boldsymbol{\tau}$ regulates the distribution of fusion weights derived from Mahalanobis distances. We conduct an ablation study by varying $\boldsymbol{\tau}$ from 1 to 50 to evaluate its influence on zero-shot generalization and downstream task performance.

As illustrated in \cref{fig:temperature}, $\text{Z}_\text{coco}$ and $\text{Avg}_\text{TA}$ exhibit a distinctive trade-off. Specifically, increasing $\boldsymbol{\tau}$ shifts the weight distribution from a highly concentrated regime toward a more uniform assignment. At lower values (e.g., $\boldsymbol{\tau} \le 10$), the sharp weight distribution prioritizes specific LoRA experts, which improves $\text{Avg}_\text{TA}$ but limits $\text{Z}_\text{coco}$. Conversely, a larger $\boldsymbol{\tau}$ promotes the aggregation of knowledge across multiple experts, yielding a steady improvement in $\text{Z}_\text{coco}$ until reaching a plateau. However, excessively high values (e.g., $\boldsymbol{\tau} > 20$) lead to pronounced parameter conflicts during the fusion process, resulting in a significant degradation of $\text{Avg}_\text{TA}$.

Empirically, the model achieves an optimal trade-off between foundational knowledge preservation and downstream task adaptation when $\boldsymbol{\tau} \in [10, 20]$. Consequently, $\boldsymbol{\tau}$ is set to 20 by default for all other experiments to ensure a robust balance across diverse scenarios.

\subsection{Stability to Task Ordering}
\label{sec:task_ordering}

To evaluate the stability of our framework, we conduct experiments across three independent runs (Seed 0, 1, and 2), with each seed representing a distinct random task ordering. The evaluation results for $\text{Z}_\text{coco}$, $\text{Avg}_\text{I}$, and $\text{Avg}_\text{TA}$ are summarized in \cref{tab:ivlod_seed}.

\begin{table*}[h]
\centering
\caption{Performance stability across varying task orders. We report results for three independent runs (Order 0, 1, 2) and their average under the IVLOD and TA-IVLOD settings.}
\label{tab:ivlod_seed}
\resizebox{\textwidth}{!}{
\begin{tabular}{c|c|ccc|ccc|ccc|ccc|ccc}
\toprule
                       \multicolumn{2}{c|}{\multirow{2}{*}{Method}}        & \multicolumn{3}{c|}{TFA} & \multicolumn{3}{c|}{iDETR} & \multicolumn{3}{c|}{ZiRa} & \multicolumn{3}{c|}{DitHub} & \multicolumn{3}{c}{ours} \\
                       \multicolumn{2}{c|}{}         & $\text{Z}_\text{coco}$  & $\text{Avg}_\text{I}$  & $\text{Avg}_\text{TA}$  & $\text{Z}_\text{coco}$   & $\text{Avg}_\text{I}$   & $\text{Avg}_\text{TA}$  & $\text{Z}_\text{coco}$   & $\text{Avg}_\text{I}$  & $\text{Avg}_\text{TA}$  & $\text{Z}_\text{coco}$   & $\text{Avg}_\text{I}$   & $\text{Avg}_\text{TA}$   & $\text{Z}_\text{coco}$   & $\text{Avg}_\text{I}$  & $\text{Avg}_\text{TA}$  \\
\midrule
\multirow{4}{*}{Order} & 0       & 45.0   & 51.9  & 23.4   & 37.6    & 56.5   & 32.3   & 45.2    & 59.0  & 35.1   & 46.6    & 65.5   & 33.2    & 46.9    & 66.6  & 64.3   \\
                       & 1       & 36.6   & 47.5  & 20.0   & 36.5    & 58.5   & 31.1   & 44.2    & 60.6  & 33.7   & 47.3    & 66.2   & 33.5    & 47.5    & 68.0  & 63.7   \\
                       & 2       & 46.3   & 51.0  & 22.7   & 38.1    & 60.2   & 38.7   & 45.8    & 62.3  & 36.1   & 46.6    & 67.3   & 30.8    & 47.2    & 68.0  & 65.4   \\
                       & average & 42.7   & 50.1  & 22.0   & 37.4    & 58.4   & 34.0   & 45.1    & 60.6  & 35.0   & 46.8    & 66.3   & 32.5    & 47.2    & 67.5  & 64.5  \\
\bottomrule
\end{tabular}
}
\end{table*}

As shown in \cref{tab:ivlod_seed}, our method demonstrates robust and consistent performance across all tested sequences. While baseline methods experience significant performance degradation and instability under the challenging TA-IVLOD setting, our approach maintains a stable $\text{Avg}_\text{TA}$ of 64.5 regardless of the task order. This consistent advantage over baselines such as DitHub and ZiRa across diverse learning paths confirms the robustness and reliability of our framework against sequential variations.

\subsection{Zero-shot Generalization on Broader Categories}
We further evaluate our method on LVIS minival\citesupp{gupta2019lvis} to examine its zero-shot generalization ability over a broader category space. Since most LVIS categories are not included in the incremental tasks, this benchmark provides a challenging evaluation of generalization to unseen concepts. As shown in \cref{tab:lvis}, our method achieves competitive performance and slightly improves over both Grounding DINO and DitHub, demonstrating that our framework effectively preserves the open-vocabulary generalization capability of the pre-trained detector.

\subsection{Efficiency Analysis}
We analyze the computational efficiency of our method under different numbers of learned classes. As shown in \cref{tab:latency}, although the inference cost increases with the number of learned classes for both DitHub and our method, the additional overhead introduced by our dynamic fusion mechanism is marginal. Specifically, our method maintains comparable latency, memory usage and the same number of learnable parameters as DitHub. This demonstrates that the proposed TTDF module improves task-agnostic detection performance without sacrificing inference efficiency.

\begin{table*}[!t]
\centering
\caption{{Efficiency analysis under different numbers of learned classes. LP denotes the number of learnable parameters.}}
\label{tab:latency}
\resizebox{0.5\textwidth}{!}{
\begin{tabular}{cccccc}
\toprule
Method                  & classes & FPS$\uparrow$ & latency$\downarrow$ & memory$\downarrow$ & LP$\downarrow$\\
\midrule
\multirow{3}{*}{DitHub} & 40      & 4.5 & 221.6ms   & 1492.7MB  &35.75M\\
                        & 30      & 4.8 & 207.3ms   & 1423.8MB &27.27M\\
                        & 20      & 5.1 & 196.8ms   & 1391.5MB  &18.79M \\
\midrule
\multirow{3}{*}{ours}   & 40      & 4.2 & 240.1ms   & 1505.1MB  &35.75M \\
                        & 30      & 4.5 & 220.8ms   & 1435.3MB  &27.27M \\
                        & 20      & 4.9 & 206.3ms   & 1401.8MB  &18.79M \\
\bottomrule
\end{tabular}
}
\end{table*}

\subsection{Analysis of Image-level Feature Weighting}
We compare our image-level TTDF with instance-level weighting based on SAM2\citesupp{ICLR2025_45c1f6a8} and UPN\citesupp{jiang2024chatrex} proposals. As shown in \cref{tab:weighting}, both instance-level variants perform worse than our image-level design. Although instance-level features may provide more localized information, their effectiveness heavily depends on proposal quality, which is often unreliable in complex multi-object scenes across different incremental tasks. In contrast, image-level features provide more stable distributional cues for expert weighting, leading to better robustness and efficiency in the TA-IVLOD setting.
\begin{table*}[!t]
\centering

\resizebox{0.3\textwidth}{!}{

}
\end{table*}

\begin{table}[!t]
\begin{minipage}[t]{0.45\linewidth}
    \centering
\caption{{Zero-shot generalization on LVIS minival.}}
\label{tab:lvis}
\resizebox{0.9\textwidth}{!}{
\begin{tabular}{cccc}
\toprule
 & G-Dino & DitHub & ours \\
\midrule
LVIS minival  & 21.1   & 21.6   & 22.0 \\
\bottomrule
\end{tabular}
}
\end{minipage}
  \hfill
\begin{minipage}[t]{0.51\linewidth}
    \centering
\caption{{Comparison between image-level and instance-level weighting strategies.}}
\label{tab:weighting}
\resizebox{0.6\textwidth}{!}{
\begin{tabular}{cccc}
\toprule
        & SAM2  & UPN  & ours \\
\midrule
$\text{Avg}_\text{TA}$ & 41.8 & 52.9 & 64.5 \\
\toprule
\end{tabular}

}
\end{minipage}

\end{table}

\subsection{Generalizability Across Models}
To further verify that our proposed modules are not restricted to a specific open-vocabulary detector, we additionally evaluate them on YOLO-World\citesupp{cheng2024yolo}. As shown in \cref{tab:yolo}, directly applying the incremental adaptation baseline to YOLO-World provides only limited improvement under the TA-IVLOD setting. In contrast, our method achieves a clear performance gain while maintaining comparable zero-shot generalization on COCO. This indicates that the proposed modules can effectively alleviate cross-task semantic interference and parameter conflicts beyond the Grounding DINO architecture. These results further demonstrate that our framework is not tied to a specific detector, but can generalize to different open-vocabulary detection backbones.

\begin{table*}[!t]
\centering
\caption{Generalization results on YOLO-World under the TA-IVLOD setting.}
\label{tab:yolo}
\resizebox{0.5\textwidth}{!}{
\setlength{\tabcolsep}{6pt} 
\begin{tabular}{cccc}
\toprule
        & YOLO-World & $+$ DitHub & $+$ ours \\
\midrule
$\text{Z}_\text{coco}$   & 46.7       & 46.3   & 46.3 \\
$\text{Avg}_\text{TA}$ & 24.4       & 25.1   & 35.6 \\ 
\bottomrule
\end{tabular}
}

\end{table*}

\section{Qualitative Results}
\label{sec:qualitative}

\begin{figure}[h]
    \centering
    \includegraphics[width=\textwidth]{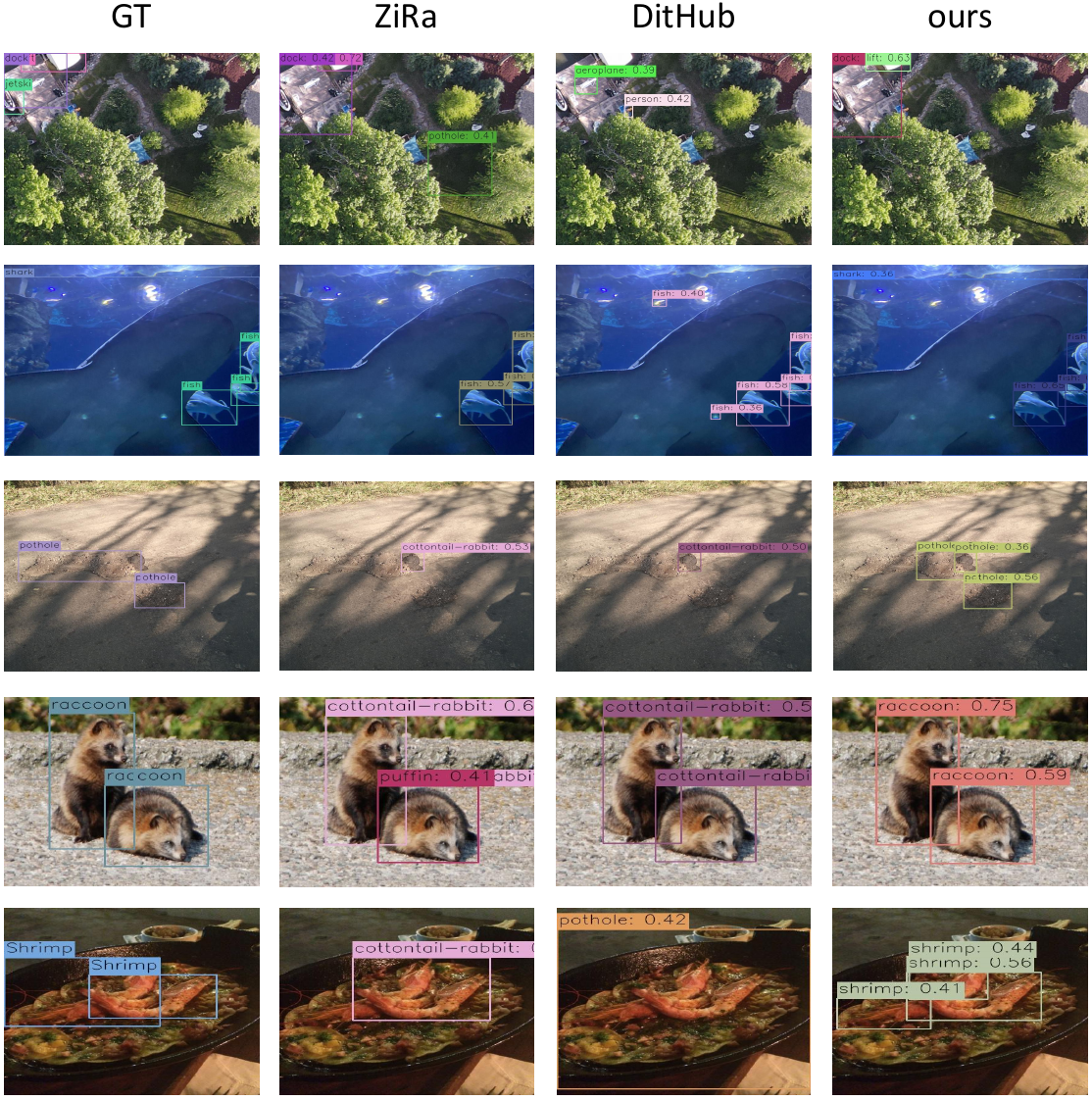}
    \caption{Qualitative comparison under the TA-IVLOD setting. Ground Truth (GT) and detection results for ZiRa, DitHub, and our method are shown.}
    \label{fig:vis_result}
\end{figure}

\Cref{fig:vis_result} presents a qualitative comparison between our proposed framework and state-of-the-art baselines (ZiRa and DitHub) under the challenging TA-IVLOD setting. When exposed to a unified prompt containing all learned categories, the baseline methods exhibit pronounced cross-task semantic interference. Specifically, they frequently misclassify objects into unrelated categories from other tasks, such as identifying raccoons, shrimps, and potholes as ``cottontail-rabbit''. In contrast, our method effectively mitigates this semantic noise. As demonstrated across diverse scenes, our approach accurately localizes and categorizes the target objects. These visualizations qualitatively confirm that our framework maintains strong discriminative capabilities and robustness against prompt-induced ambiguities.

\bibliographystylesupp{splncs04}
\bibliographysupp{main}

\end{document}